\documentclass[conference]{IEEEtran}

\usepackage{cite}
\usepackage{amsmath,amssymb,amsfonts}
\usepackage{algorithmic}
\usepackage{subcaption}
\usepackage{graphicx}
\usepackage{textcomp}
\usepackage{xcolor}
\usepackage{float} 
\usepackage[font={small}]{caption}

\newcommand{\eqnref}[1]{(\ref{eqn:#1})}

\newcommand{\eqnlabel}[1]{\label{eqn:#1}}

\newcommand     {\paren}[1]{\left(#1\right)}

\newcommand{\curlb}[1]{\left\{#1\right\}}
\newcommand{\squareb}[1]{\left[#1\right]}
\newcommand{\eX}[1]{e^{#1}}

\newcommand{\norm}[1]{\left\|#1\right\|}
    
\begin{document}

\title{\Large A look at adversarial attacks on radio waveforms from discrete latent space
}
\author{
\IEEEauthorblockN{\normalsize Attanasia Garuso, Silvija Kokalj-Filipovic, Yagna Kaasaragadda}
\IEEEauthorblockA{\small Rowan University\\
\small\em garuso38@students.rowan.edu, \{kokaljfilipovic,kaasar57\}@rowan.edu}}

\maketitle
\begin{abstract}
Having designed a VQVAE that maps digital radio waveforms into discrete latent space, and yields a perfectly classifiable reconstruction of the original data, we here analyze the attack suppressing properties of VQVAE when an adversarial attack is performed on high-SNR radio-frequency (RF) datapoints. To target amplitude modulations from a subset of digitally modulated waveform classes, we first create adversarial attacks that   preserve the phase between the in-phase and quadrature component whose values are adversarially changed. We compare them with adversarial attacks of the same intensity where phase is not preserved. We test the classification accuracy of such adversarial examples on a classifier trained to deliver 100\% accuracy on the original data. To assess the ability of VQVAE to suppress the strength of the attack, we evaluate the classifier accuracy on the reconstructions by VQVAE of the adversarial datapoints and show that VQVAE substantially decreases the effectiveness of the attack.
We also compare the I/Q plane diagram of the attacked data, their reconstructions and the original data. Finally, using multiple methods and metrics, we compare the probability distribution of the VQVAE latent space with and without attack. Varying the attack strength, we observe interesting properties of the discrete space, which may help detect the attacks.  %As the same VQVAE can be trained to denoise the channel effects, we propose to test the effectiveness of VQVAE under simultaneous channel and adversarial perturbations, which we omit here due to space constraints.  
\end{abstract}
\vspace{-2mm}
\section{Introduction}
Breakthroughs in AI have motivated their adoption in NextG standards and practices. In addition, radio-frequency (RF) signal processing based on machine learning (RFML) has already made significant contributions to spectrum sensing, and it continuously motivates new research efforts. Baseband physical layer signal processing which was traditionally  performed in centralized baseband units (BBUs) of cloud radio access networks (C-RAN) is now  AI-empowered and is being distributed across many Open-RAN components. While Open-RAN may better address the communication overhead, service heterogeneity, and computations needed to implement NextG RAN functions, the NextG AI algorithms will require substantial data transfer between its components, even on the level of baseband signals. This calls for careful consideration of data vulnerabilities. Further, although end-to-end AI-native communication systems may revolutionize how data is embedded in radio signals, baseband data as the foundation of  communication protocols will certainly prevail at least as one of the information carriers.  This includes different semantic communication frameworks that are being proposed, sharing a common property of heavy reliance on AI.
Although some novel semantic communication frameworks propose a
nalog communication over wireless channels, it is likely that semantic communications will rely on hybrid analog-digital schemes or the traditional digital communication concepts. Digital communication is likely to persist, for the reasons of legacy, robustness and security. Tying to it the fact that NextG heavily relies on deep learning, which can be deceived by RF adversarial data, makes it obvious that adversarial attacks on digitally modulated signals still deserve the attention of the research community. RFML requires massive training data.  Data augmentation using AI is the key to providing sufficient data for diverse AI algorithms supporting NextG networks \cite{reformer}. The sheer amount of RF data for training, including the augmented data, represents a massive attack space. Moreover, the landscape is shifting, and new attack types are likely to surface. E.g., the data transmitted over a wireless channel in semantic communications will bootstrap semantic reconstructions affecting much wider domains than mere source-decoding, hence the effects of adversarial attacks may be more dangerous.  

There has been plenty of research on the topic of RF adversarial attacks in RF \cite{Adexdetect, AEagainst, sadeghi2018adversarial,realizationRFadvers}.  We here direct our attention to how a vector quantized variational-autoencoder (VQVAE) can be applied to attacked RF samples, received or stored by the victim, to mitigate or detect the attack. 
Although autoencoders have been previously used to mitigate RF adversarial examples (AdExs) \cite{AEagainst}, discrete-domain autoencoders have not been studied as mitigation mechanisms. Attack detection has also been one of the research topics \cite{Adexdetect, lidDetect, nullFeatAdex}: various statistical methods are applied to identify AdExs in order to isolate them. 
If the statistical test is based on the distribution shift between the original data and the adversarial examples, such a test will become unusable if the original data experiences a shift independent from the adversarial attack. The multipath effects in an over-the-air attack would cause such a shift. The comprehensive analytical model in \cite{realizationRFadvers}, which includes channel effects, is based on multiple assumptions, including perfect synchronization and perfect knowledge of the channel, both between the adversary and the receivers, and between the transmitter and the receivers. Despite the fact that these assumptions are very difficult for an adversary to fulfill, the work in \cite{ realizationRFadvers} proves that over-the-air RF adversarial attacks are realizable. 
Although we studied the attacks on a classifier that has been trained on either the clean or multipath-affected RF signals, which emulates the over-the-air attack from \cite{ realizationRFadvers}, this paper only presents attacks on the clean data, due to space limits. The clean data comes from high-SNR signals, obtained after channel estimation and equalization. This emulates the case where baseband data is moved to Open-RAN components for signal processing or AI training, or stored at the edge, and is therefore exposed to an attack off-the-air. New data-hungry AI-based channel models \cite{MIMOChannelEstimScore,  ScoreDiffChannel, yu2024ai, zhou2024generativediffusionmodelshigh}, which demonstrated outstanding performance, would be negatively affected by such data poisoning. Hence, the mitigation by VQVAE demonstrated in this paper is an important contribution. Section~\ref{sec:system} introduces
the problem and describes data, attacks and the VQVAE model. Section~\ref{sec:eval} describes  methods of
evaluation. Section~\ref{sec:res} discusses the results of the evaluation, followed by the Conclusion section. 
%%%%%%%%%%%%%%%
\vspace{-1mm}
\section{Problem Definition}\label{sec:system}
We train a classifier using the original dataset $X,$ composed of 6 radio waveform classes. 
%The weights of the trained classifier are represented by $\Theta_c$. 
Adversarial attacks on that classifier were generated using two primary methods:  FGSM \cite{szegedy2013intriguing} and PGD \cite{madry2017towards} attacks. For FGSM, we implemented two variations: FGSM1, a phase preserving attack and FGSM2, typical FGSM attack which creates independent perturbations on the in-phase~(I) and quadrature~(Q) components.
For some of the waveform modulations in $X$, the phase component carries  critical information  about the signal characteristics and embedded information.  Altering the phase  extremely  can lead to easily detectable AdExs and undecodable information. We expect these modulations to be less vulnerable to FGSM1. Amplitude modulations embed the information bits by modifying the amplitude according to a discrete mapping, hence they are expected to be vulnerable to both FGSM1 and FGSM2.

AdExs from all three attacks (FGSM1,FGSM2 and PGD) are classified by the classifier and the accuracy metric is calculated. The same AdExs $X^{af1}, X^{af2}, X^{ap}$ are then passed through a VQVAE trained on $X.$ VQVAE outputs both the discrete latent representation $Z_q$ of the input $x$ and a reconstruction $\hat{x}$ of $x$ based on the latent $Z_q$. We collect reconstructions $\hat{X}^{af1}, \hat{X}^{af2}, \hat{X}^{ap}$ and the latent mappings $Z_q^{af1}, Z_q^{af2}, Z_q^{ap}.$ The adversarial reconstructions will be tested on the classifier to demonstrate the VQVAE ability to mitigate the adversarial attack. We also analyze latent mappings from adversarial examples of different strength. We next describe the original dataset $X$.
\vspace{-1mm}
\subsection{Dataset}\label{subsec:torchsigdata}
%\vspace{-1mm}
Our dataset consists of datapoints which represent sampled RF signals, of the types common in modern wireless communications. Each datapoint arises from a specifically modulated sequence of random information bits, converted to a baseband RF signal. The datapoint $x$ can be represented as $x=\squareb{Re_i+j Im_i}, i = 1\cdots n$ with $j=\sqrt{-1}$.
Here, a modulated signal
$u$ is obtained as $u = M_s(b)$, where $s \in \mathcal{S}$ is the employed
modulation scheme, with $S$ denoting the finite set of available
digital modulation schemes. In this paper, we extend the notion of modulation, as we included the OFDM waveform, hence $$\mathcal{S} = \squareb{\text{4ask,8pam,16psk,32qam-cross,2fsk,ofdm256}}.$$ For any $s$, $M_s = \curlb{0,1}^m \rightarrow \mathcal{C}^v$ describes the modulation function of modulation class $s$. The random sequence of bits $b=\curlb{0,1}^m$ of length $m$ is encoded into a sequence of complex valued numbers of length $v,$ where the complex sample $c_i=Re_i +j Im_i,1 \leq i \leq v$ encodes the modulation phase $\phi=\arctan{Re_i/Im_i},$ and amplitude $a_i = \sqrt{Re_i^2+Im_i^2}.$ We create datapoints as sub-sequences  $x$ of $u \in \mathcal{C}_v,$ of length $n = 1024.$ We prepared the training dataset $X_{train}$ by using an open-source library {\em torchsig} featured in \cite{torchsig}. $X_{train}$ contains RF samples of high SNR, for both simplicity and  future utility in creating various signal augmentations controlled by a prompt. The torchsig library function {\em ComplexTo2D} is used to transform vectors of complex-valued numbers into  2-channel datapoints, traditionally referred to as {\em I} and {\em Q} components. Each channel is comprised of $n$ real numbers, previously normalized. Channel 1 contains real components $I\in \mathbb{R}^n$ and channel 2 contains the imaginary ones $Q\in \mathbb{R}^n.$  
$\mathcal{S}$ contains 2 amplitude modulations (PAM, ASK). PAM is of interest as it is highlighted in low-complexity, power-efficient, and short-range mmWave backhaul applications \cite{saad2020back, backhaul, mmwbackhaul}, or in higher frequency (e.g., sub-THz) extensions for future wireless systems. While both PAM and ASK are analog modulations, PAM uses distinct amplitude levels to represent data, whereas ASK uses different phases of the carrier to represent data. $\mathcal{S}$ includes 3 phase-based modulation classes: 16-PSK, 32-QAM-X, and OFDM256, whose carriers are modulated by 256-QAM. QAM modulations can also be considered both amplitude and phase modulations. Finally, binary FSK (2FSK) and 16-PSK are constant-envelope modulations.  Every modulation contributes with datapoints created by sampling the waveform with 2 samples per symbol, except for 2FSK
signal which is oversampled at 8 IQ samples per symbol to avoid aliasing distortion \cite{torchsig}.
%\addtolength{\topmargin}{+0.03cm}
%%%%%%%%%%%%%%%%%%%%.
\begin{figure}[h]
\centering
%\vspace{+1mm}
\includegraphics[width=0.5\textwidth, height = 4.5cm]{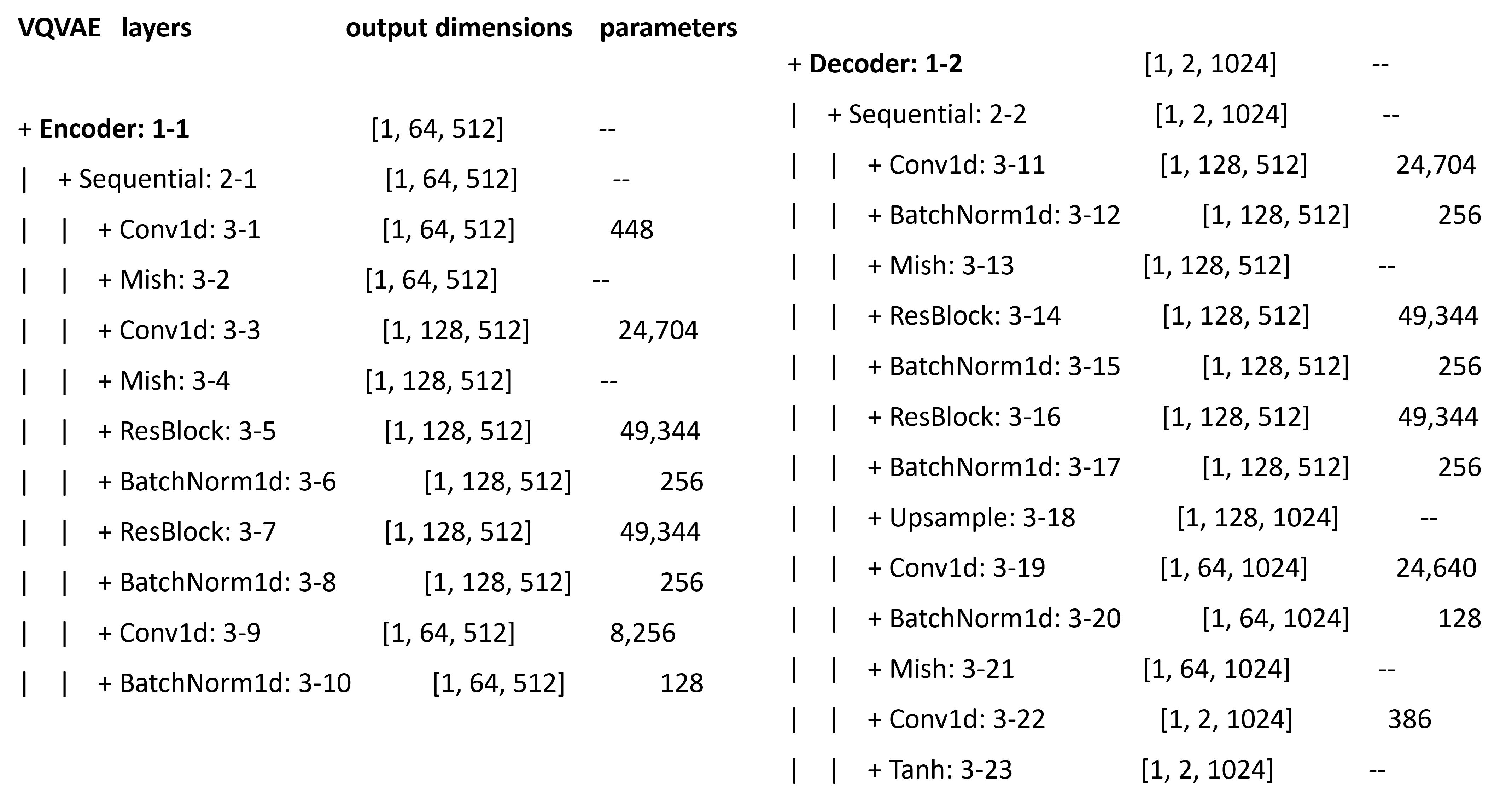}
\vspace{-3mm}
\caption{{\bf VQVAE's compact architecture: output dimensions for a unit batch, and \# of trainable parameters per layer.}
}\vspace{-3mm}
\label{fig:vqvaearch}
\end{figure}
%%%%%%%%%%%%%%%%%%%%.
\vspace{-1mm}
\subsection{Attack Methods}
For the attack known as Fast Gradient Sign Method (FGSM), renamed here as FGSM2,  a classification model with weights $\Theta_c$ is attacked by modifying the input data $x$ imperceptibly while maximizing the probability of misclassification. This optimized attack leverages the knowledge of the loss function calculated by the model by adding a perturbation of strength $\epsilon$ in the direction of the loss gradient (increasing the loss): $x_a =x+\epsilon\ sign(\Delta_x J(\Theta_c,x,y)).$ $J$ is the cross-entropy loss function evaluated for input $x$ of label $y.$ For our $x,$ composed of 2 channels (I and Q) with 1024 samples each, the attack is performed independently on each channel: $I(Q)_a =I(Q)+\epsilon \ sign(\Delta_{I(Q)}J(\Theta_c,I(Q),y)).$
For PGD, the same  attack is performed over iterative steps, and then $x_a$ is projected back into the ball of radius $\epsilon$ around $x$ to ensure the adversarial example remains within the constraints of the attack strength. Because of the iterative search, PGD is more effective.

For FGSM1, i.e. the FGSM phase-preserving attack, we first record the phase of the original complex number corresponding to the 2-channel $x$ as $\varpi=arctan(Q,I)$. After independently adding perturbations to I and Q, we calculate the amplitude of the resulting complex number
$A_a=\sqrt{I^2_a+Q^2_a}$, and then calculate the final AdEx channel components as
\begin{align}
\vspace{-3mm}
&I_a=A_a\cos{\varpi} \\
&Q_a=A_a\sin{\varpi}. \label{eq:FGSM1adex}
\end{align}
%%%%%%%%%%%%%%%%%%%%%%%%%%%%%%%%%%%%
\subsection{VQVAE}
Prior work \cite{reformer} introduced a similar VQVAE architecture to generate discrete tokens for training transformer-based generative models of RF datapoints. Specifically, VQVAE has been trained to learn conditional discrete distribution $P(Z_q|x)$, while the transformer is trained to learn $P(Z_q)$ and generate fake discrete latents. The VQVAE's decoder then maps the fake latents back to the signal space. This computationally efficient procedure was proposed  instead of learning the prior $P(x),$ which would allow us to sample $P(x)$ and generate similar RF fakes, only {\em with much higher computational cost}.

Using a vector-quantized variational autoencoder model \cite{vqvae2017neural} allows us to learn not only an
efficient, discrete, low-dimensional 
representation $Z_q$ of every RF datapoint $x,$ but also a discrete mapping $E_q:X\rightarrow \mathbb{Z}^{q_s}.$ Here, $E_q$ is a combination of an encoder and vector-quantizer, $q_s$ is the size of the codebook $Q,$ $Z_q$ is the vector-quantized version of the latent $Z_e,$ the output of the VQVAE encoder $E$ given the input $x$.  
 Using the mapping $E_q(X_{train}),$ composed of $Z=E_\Theta(x)$ and $\Omega_Q(Z),$ the  vector quantizer based on the trained codebook $Q$, the dataset $ZD$ is mapped from the training dataset $X_{train}.$ Equations~\eqnref{vqvaefwd}  represent the trained VQVAE model utilized for the above mapping.
%%%%%%%%%%%%%%%%%%%%
\begin{align}
\vspace{-2mm}
\nonumber Z_e &= E(x, \theta_E) \\
\nonumber Z_q &= \Omega(Z_e, Q, \theta_Q) \\
    \hat{x} &= D(Z_q, \theta_D), \eqnlabel{vqvaefwd}
\end{align}
%%%%%%%%%%%%%%%%%%%%
where $\theta_E,$ $\theta_Q,$ and $\theta_D$ are the trained weights of the encoder, codebook and decoder, resp.
For the results presented in this paper, $q_s=128$ codewords of length $\ell=512.$ Every $\ell$-element slice $z_i$ of the $Z_e$ is stochastically quantized to a codeword in $Q.$ Please see the architecture of VQVAE, which includes dimensions of $Z_e$ at the output of $E_{\theta_E}$ in Fig.~\ref{fig:vqvaearch}. This means that $Z_q$ will be a vector of 64 integers between $1$ and $128.$ Stochastic vector quantization is used as a regularizer to prevent the infamous mode collapse \cite{huh2023straightening}. We here refer to such stochastic mapping as $\Omega_{SC}.$ While mapping the slices of $Z_e$ to the codebook, VQVAE model is simultaneously training the codebook vectors. The decoder block, denoted $ D(Z_q) $, is tasked with reconstructing the original input $x$ from the mapped codebook vectors.  
%The architecture of VQVAE is presented in Fig.~\ref{fig:vqvaearch}.
VQVAE is trained on the RF samples with a high SNR and without wireless channel effects. One of the training objectives is for 
the reconstructed output, referred to as $\hat{x}$, 
to be as close as possible to the input (or 
the original) $x$, expressed by the reconstruction loss:
$\mathcal{L}_{rec}=\frac{1}{p}\sum_{i=1}^{p}{\paren{x_i-\hat{x}_i}^2}.$
%%%%
Note that we apply stochastic quantization $\Omega_{SC}$ by sampling the codebook \cite{takida2022sq} according to the learned discrete posterior
\begin{align}
P(Z_Q[i] = k\,|\,x) &= \eX{-\norm{z_i(x) - e_k}^2} 
\eqnlabel{post}
\end{align}
with $k$ denoting the index of the codeword $e_k,$ and $z_i(x)$ denoting a slice of $Z_e.$
Hence, the loss also includes the KL divergence between the posterior in \eqnref{post}  and the discrete uniform prior $P_d(k)=1/q_s$, $KL(P(k|x)||P_d(q))$. To properly learn the latent space $Z_q$ and $P(Z_q|x),$ loss function is expanded with 
\begin{equation}
\mathcal{L}_{total}=\mathcal{L}_{rec}+\mathcal{L}_{quant}+\beta\paren{\mathcal{L}_{commit}+KL}.
\end{equation}
In addition, the poorely utilized codewords are reset during the training. 
For more detail about the design and training of VQVAE, please see \cite{reformer}.
\vspace{-2mm}
\subsection{Classifier}\label{subsec:class}
\begin{figure}[h]
%\vspace{-2mm}
\centering
\hspace{-1mm}\includegraphics[width=0.48\textwidth, height = 4.4cm]{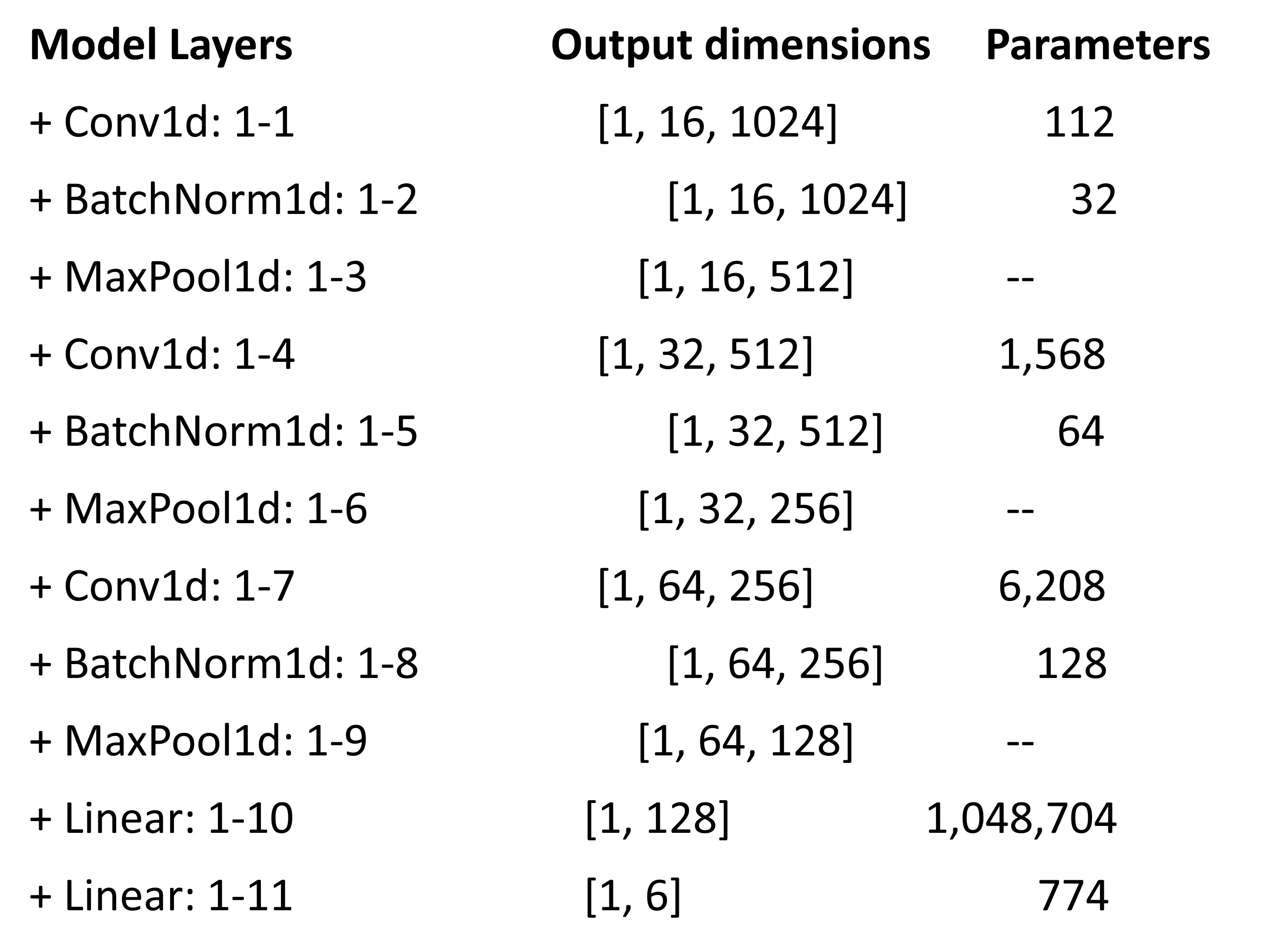}
\vspace{-1mm}
\caption{{\bf Classifier architecture with the output dimension and \# of trainable parameters per layer.}
}\vspace{-3mm}
\label{fig:classarch}
\end{figure}

The classifier trained on $X_train$ and evaluated on $X, X^{af1}, X^{af2}, X^{ap}, \hat{X}^{af1}, \hat{X}^{af2}, \hat{X}^{ap}$, has a simple architecture described in Fig.~\ref{fig:classarch}.
%%%%%%%%%%%%%%%%%%%%
%%%%%%%%%%%%%%%%%%%%%%%%%%
\section{Evaluation Methods}\label{sec:eval}
\begin{figure}[h]
%\vspace{-3mm}
\centering
\includegraphics[width=0.4\textwidth,height=3.5cm]{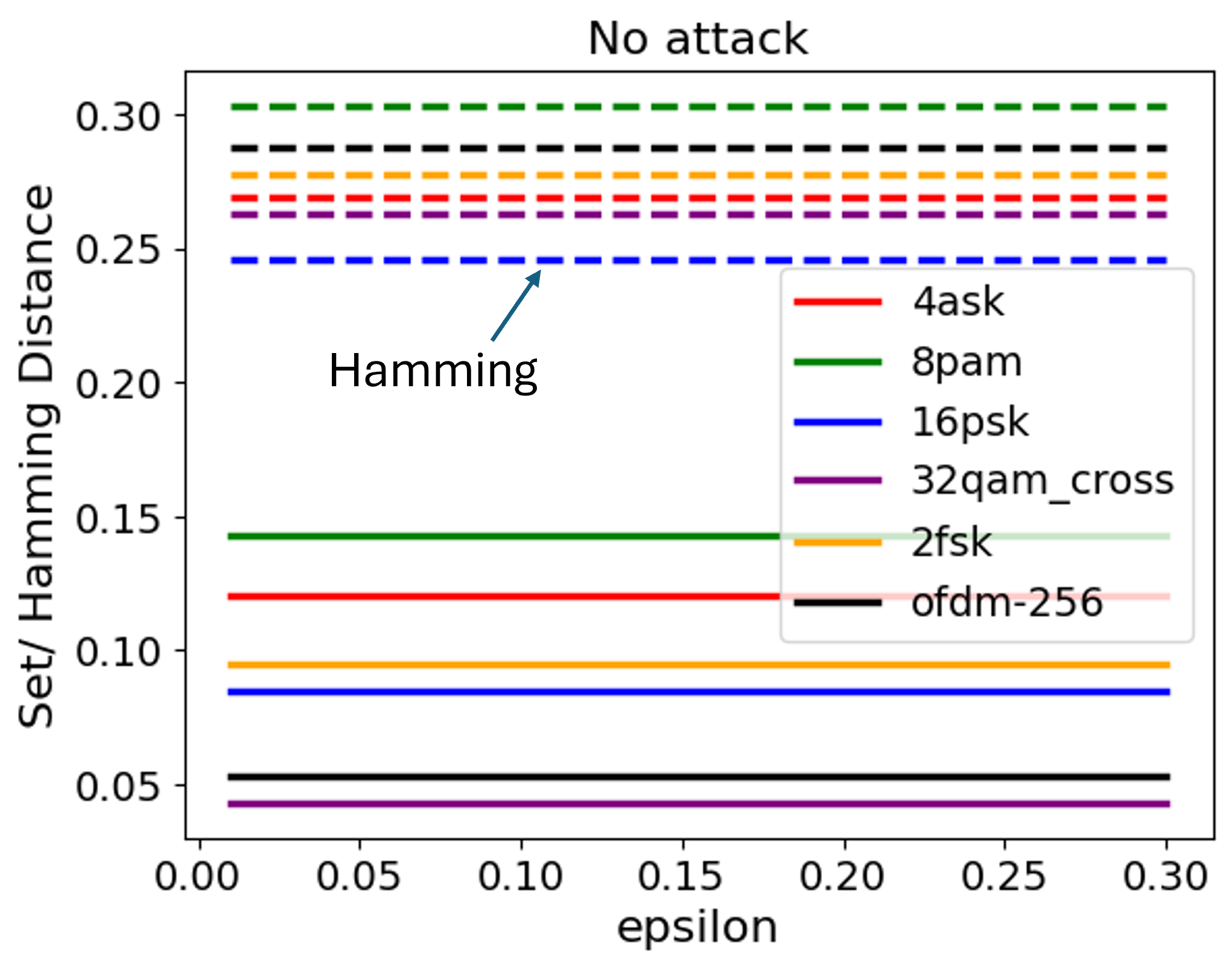} % Replace with actual file path
\caption{{\bf Average Set and Hamming distances without attack, due to $\Omega_{SC}$ (between $\mathbf{z_q}^{(i)}(x)$ and $\mathbf{z_q}^{(ii)}(x)$}}. \vspace{-4mm} 
\label{fig:residualDist}
\end{figure}

As described in subsection~\ref{subsec:class}, to evaluate the effectiveness of VQVAE-based mitigation, we compare the classification accuracy of the originals, AdExs and AdEx reconstructions by VQVAE. We do it for all 3 attack types, using multiple attack strengths in each:  $\epsilon \in \curlb{0.01, 0.06, 0.1, 0.2, 0.3}.$ %We are also comparing standard deviations of the original dataset [TODO], the attacked one and their reconstructions. Finally, we are calculating the fidelity and diversity metrics of adversarial datasets and their reconstructions with respect to the original data [TODO]. 
To demonstrate the vulnerabilities per waveform class, we also present confusion matrices. For different attack types and different $\epsilon$  we create histograms of the latents $Z_q^{af1}, Z_q^{af2}, Z_q^{ap}$ produced by VQVAE from generated AdExs, and compare them with histograms of the original latents $Z_q$. Apart from this statistical comparison, we seek to find how the discrete mapping for each single datapoint $x$ is affected by the adversarial attack by looking at the average {\em Hamming distance} $d_H$ between $Z_q(x)$ and $Z_q(x_a).$ As the performed vector quantization is stochastic, i.e. the same $x$ will produce slightly different $Z_q$ in different evaluations of $E_q(x)\sim P(Z_q|x),$ the average Hamming distance is non zero for any 2 independent drawings $\mathbf{z_q}^{(i)}\sim P(Z_q|x)$ and $\mathbf{z_q}^{(ii)}\sim P(Z_q|x)$ (Fig.~\ref{fig:residualDist}). For this reason, we normalize the average Hamming distance $d_H(Z_q(x),Z_q(x_a))$ with average $d_H(\mathbf{z_q}^{(i)}(x),\mathbf{z_q}^{(ii)}(x)),$ to obtain a metric that quantifies the impact of the attack only (Fig.~\ref{fig:Distances}).

Lastly, we compute the {\em Set distance} $d_S$ as the average size of the set-difference between the indices in $Z_q(x)$ and the indices in $Z_q(x_a),$ normalized by the size of the set-difference between the indices in 2 consecutive drawings of $E_q(x).$ Distances $d_S(\mathbf{z_q}^{(i)}(x),\mathbf{z_q}^{(ii)}(x))$ and $d_H(\mathbf{z_q}^{(i)}(x),\mathbf{z_q}^{(ii)}(x))$ are independent of the attack, illustrating the natural randomness of the applied quantization (see Fig.~\ref{fig:residualDist}).  
While $d_H(Z_q(x),Z_q(x_a))$ counts the number of positions in $Z_q,$ modified by the attack, $d_S$ is the normalized volume of the difference between sets of codeword indices in $Z_q(x)$ and $Z_q(x_a)$.
%%%%%%%%%%%%%%%%%%%%%
\vspace{-2mm}
\section{Evaluation Results}\label{sec:res}
\begin{figure}[h]
\centering
\vspace{-2mm}
\includegraphics[width=0.45\textwidth,height=4.0cm]{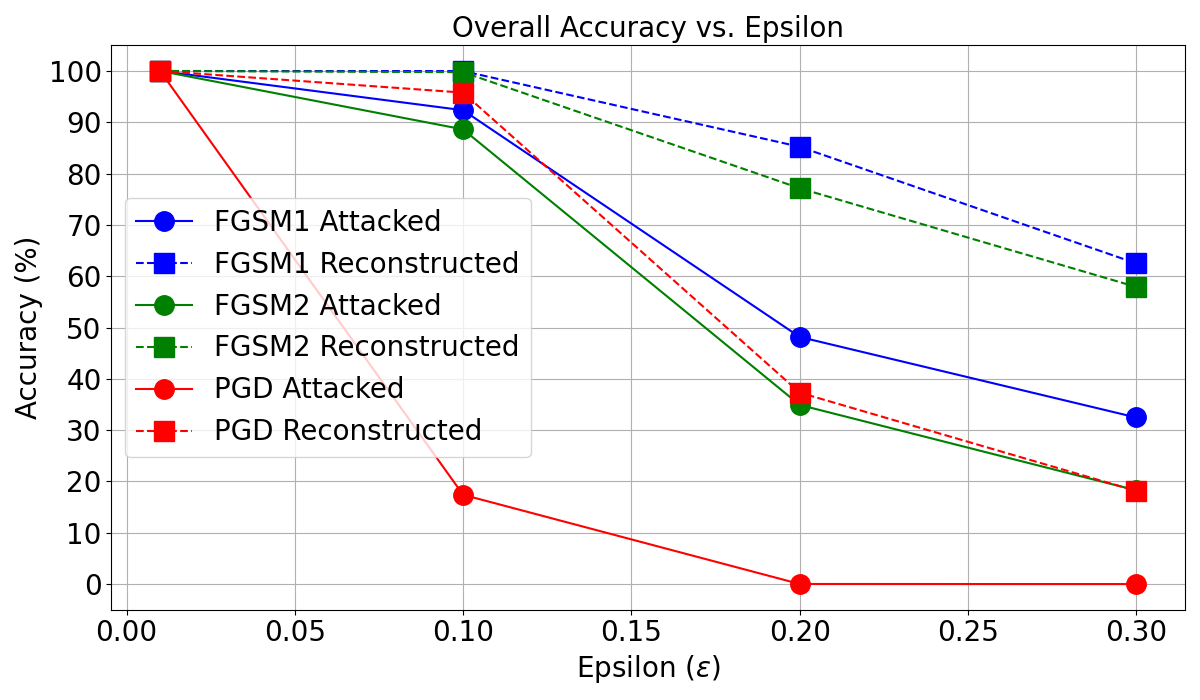} % Replace with actual file path
\caption{{\bf Accuracy under attacks}}\vspace{-4mm}
\label{fig:accuracy}
\end{figure}

\begin{figure}[h]
\centering
\begin{tabular}{c}   
 \vspace{-2mm}
 \hspace{-3mm} \includegraphics[width=0.45\textwidth,height=7.5cm]{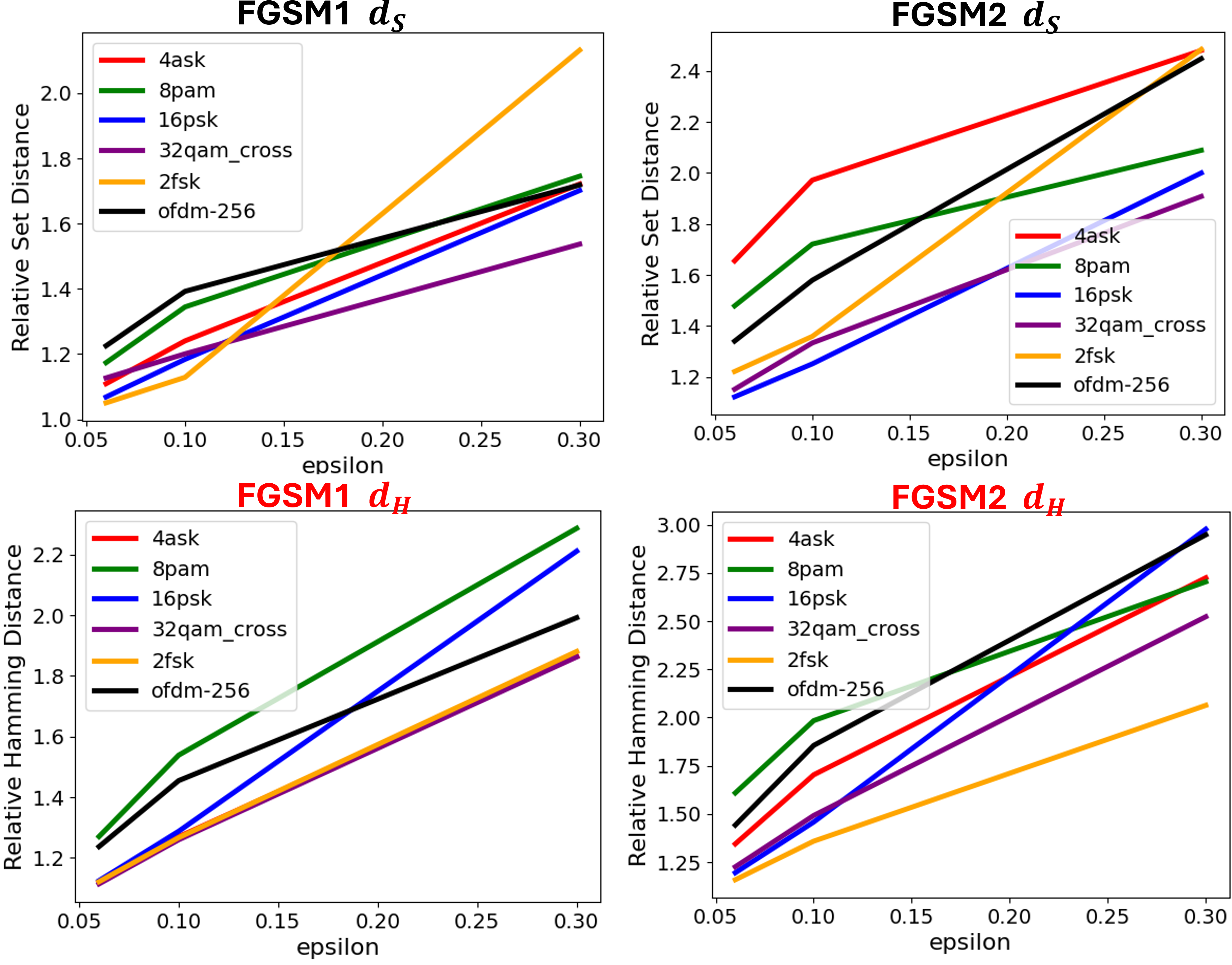} \\
 \hspace{-3mm} \includegraphics[width=0.45\textwidth,height=7.5cm]{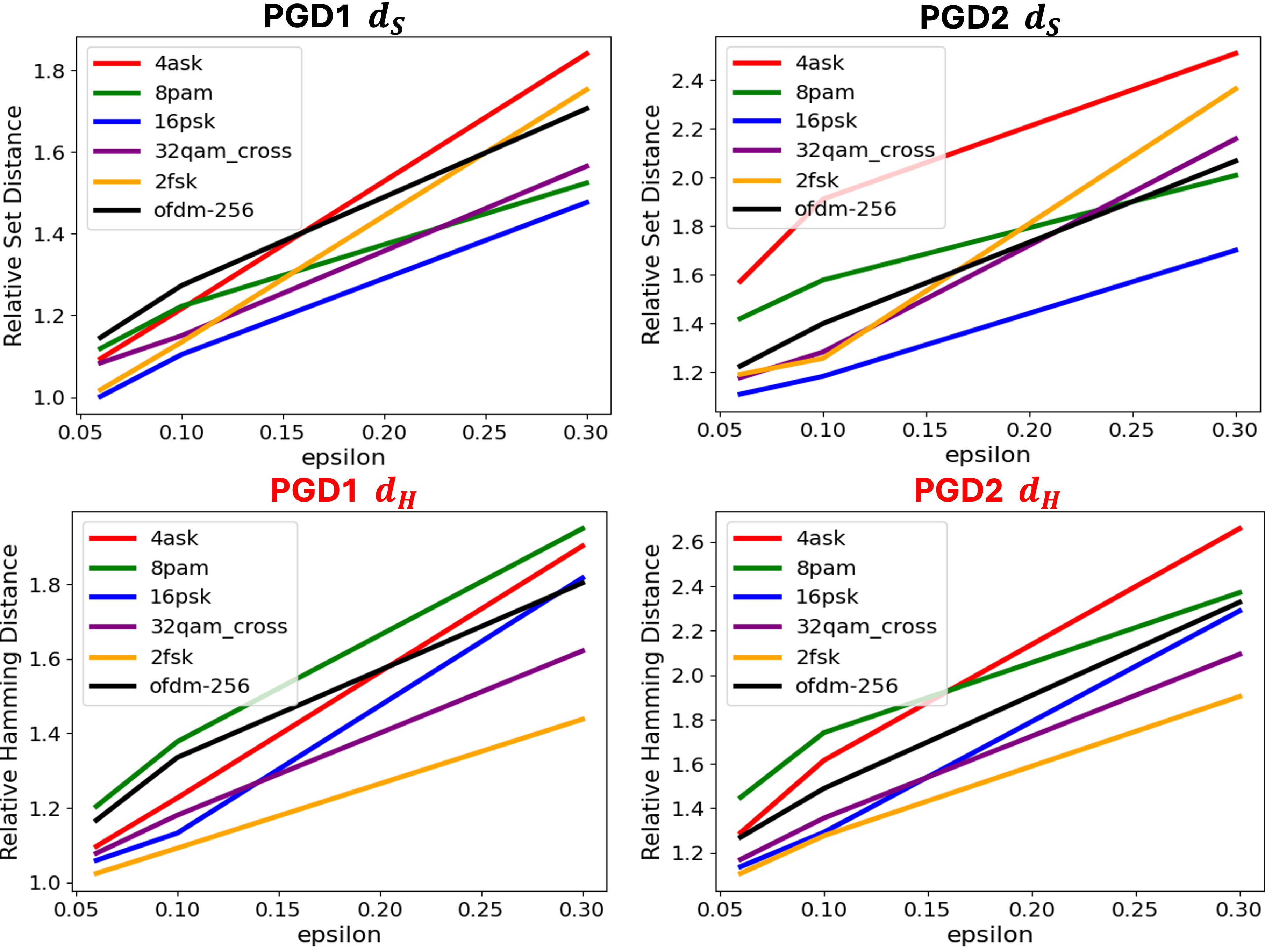} 
\end{tabular}
\caption{{\bf Hamming $d_H$ and Set distances  $d_S$ under attacks.}}\vspace{-6mm} 
\label{fig:Distances}
\end{figure}

\begin{figure}[h]
\centering
\begin{tabular}{c}   
 \hspace{-3mm} \includegraphics[width=0.45\textwidth,height=2.2cm]{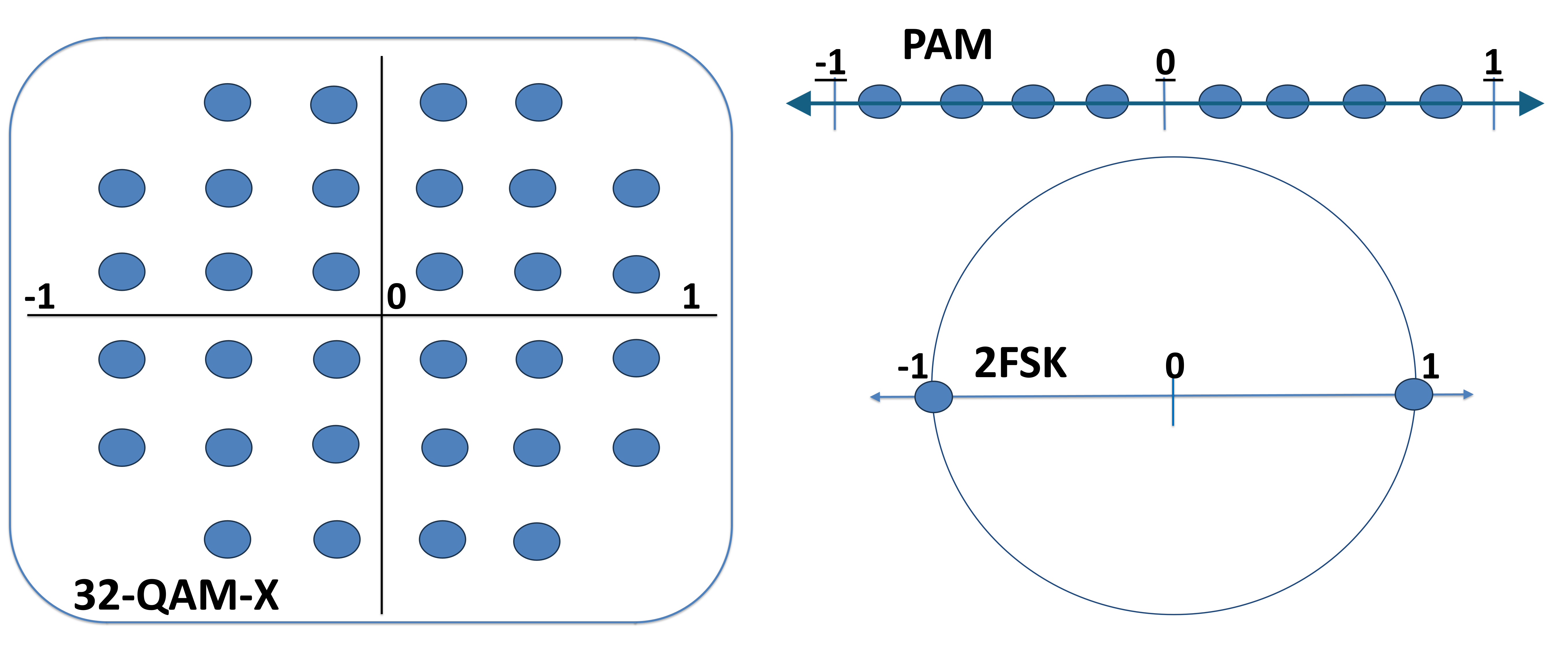} \\
 \hspace{-3mm} \includegraphics[width=0.45\textwidth,height=2.0cm]{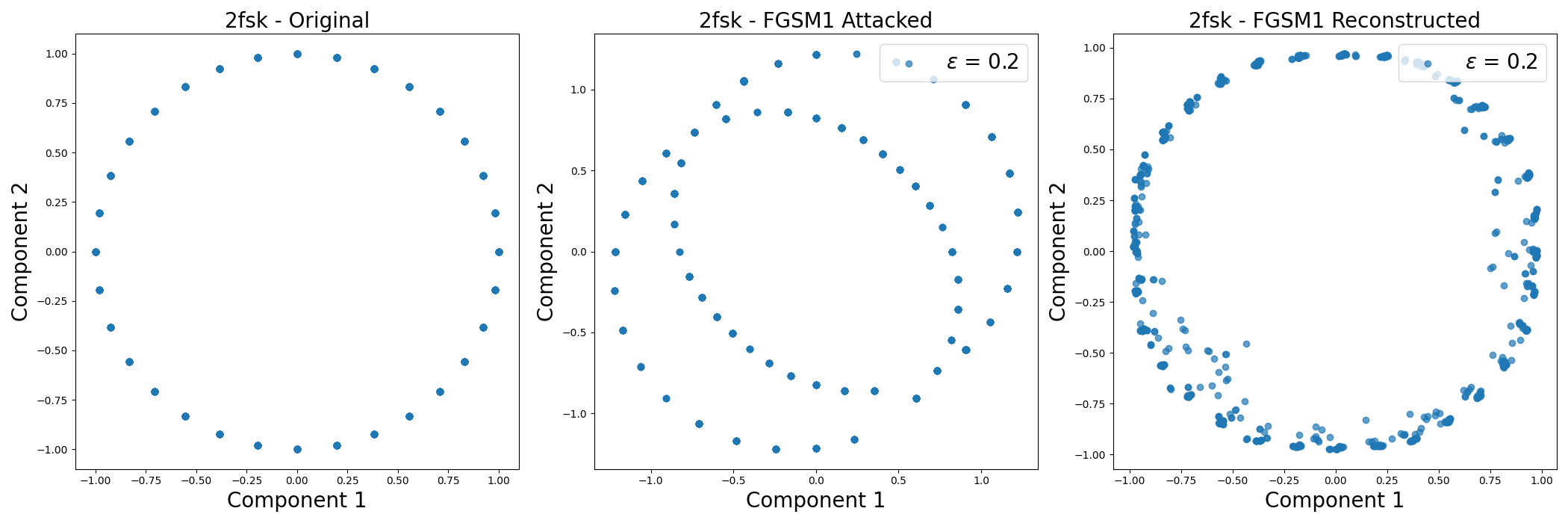}\\
 \hspace{-3mm} \includegraphics[width=0.45\textwidth,height=2.0cm]{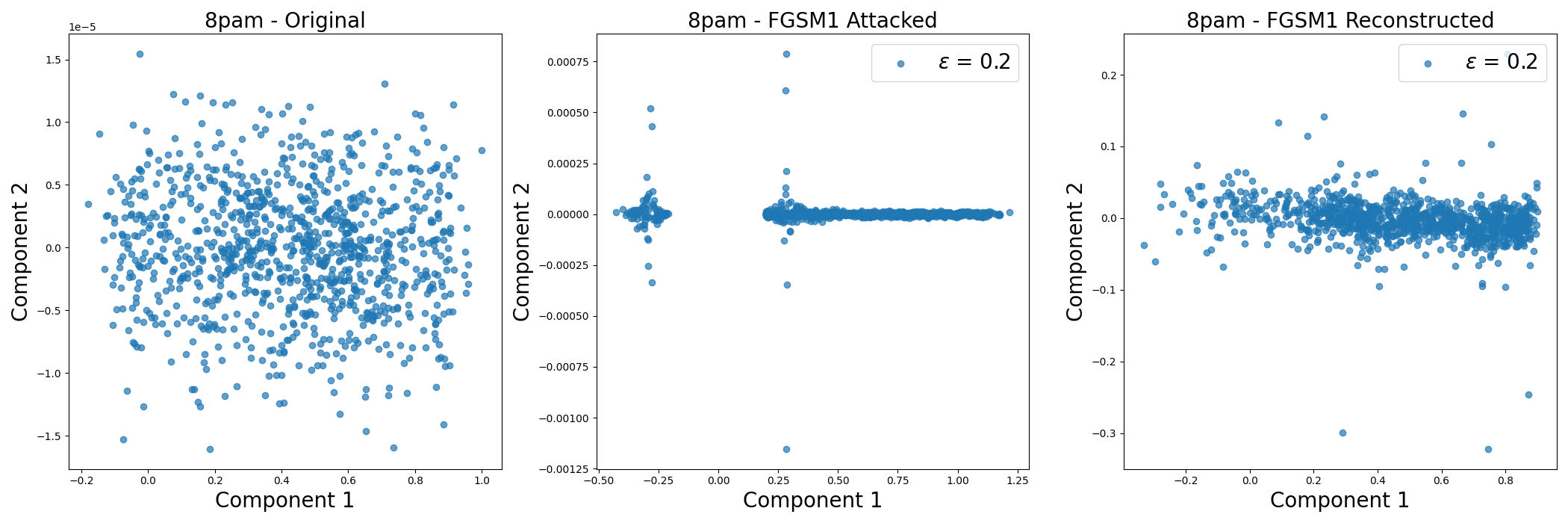}\\
 \hspace{-3mm} \includegraphics[width=0.45\textwidth,height=2.0cm]{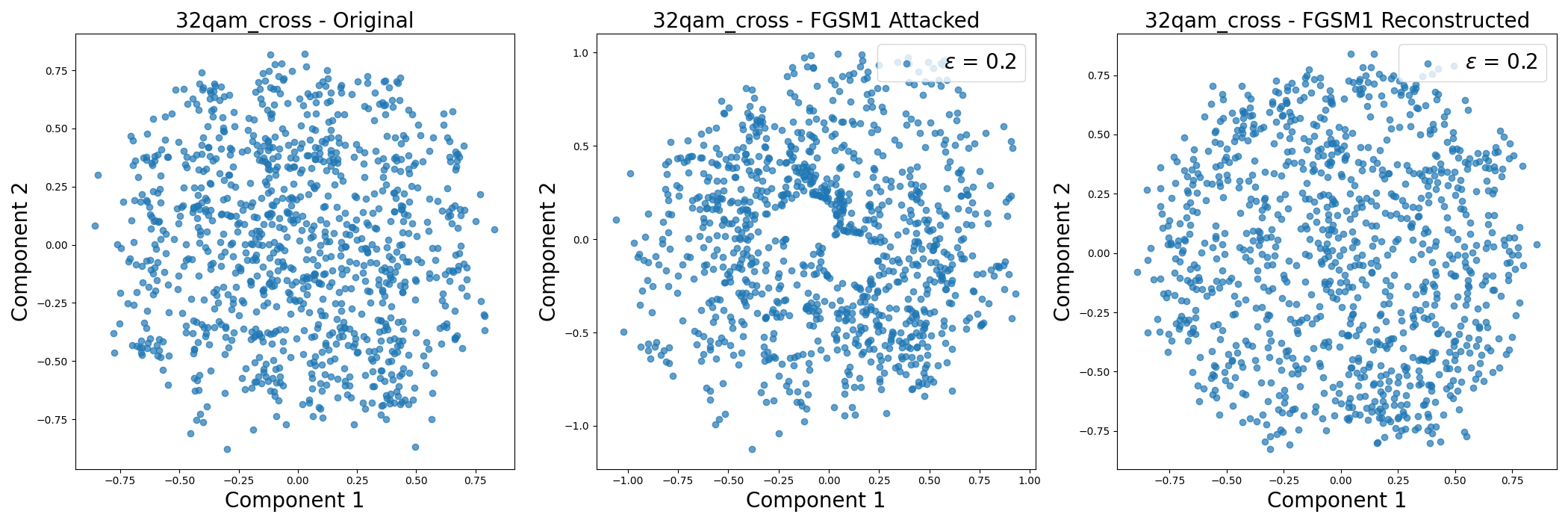}
\end{tabular}
\caption{{\bf Constellations of 2FSK, 8PAM, 32-QAM-X, and matching I/Q planes for a random datapoint of each class. }}\vspace{-3mm}
\label{fig:constells}
\end{figure}

\begin{figure}[h]
\centering
%\vspace{-2mm}
 \hspace{-3mm} \includegraphics[width=0.45\textwidth,height=7.0cm]{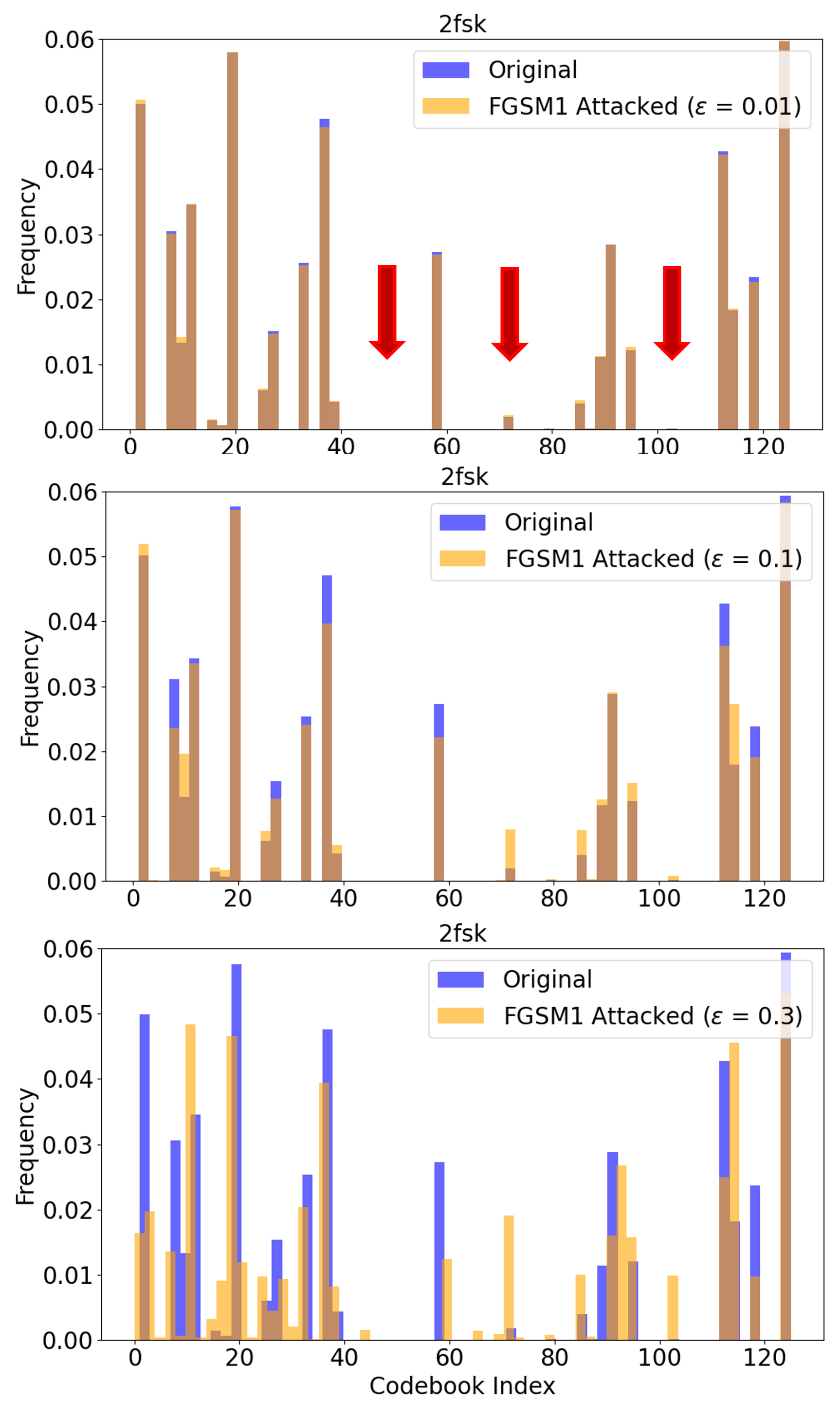} 
\caption{{\bf 2FSK histogram of codeword indices under FGSM1 (0.01, 0.1 and 0.3). The arrows show the ranges outside of the support of normal latents, which get populated by the attack, thus increasing Set distance $d_S.$}}\vspace{-5mm}
\label{fig:FSKhist}
\end{figure}

Due to space limitations, the results of the above methods are presented only for FGSM1, with the exception of  accuracy in Fig.\ref{fig:accuracy} 
 and Hamming/Set distances in Fig.~\ref{fig:Distances}. 
Fig.\ref{fig:accuracy} presents the accuracy of the classifier under 3 types of attack (FGSM1, FGSM2 and PGD(2)), and for the same attacked datapoints when they are reconstructed by VQVAE (with square markers). PGD1 is omitted as it behaves similarly with respect to PGD2, as FGSM1 wrt FGSM2. It is obvious that VQVAE helps mitigate the attack, the more so for stronger attacks ($\epsilon\geq 0.2$). To gain insight into how VQVAE is helping, we first look at the confusion matrices and then at the metrics we derived from the latent space. 
%%%%%%%%%%%%%%%%%%%%%%%%%%%%%%%%%%%%%%%
%%%%%%%%%%%%%%%%%%%%%%%%%%%%%%%%%%
\begin{figure}[h]
\centering
\vspace{1mm}
\hspace{-3mm} \includegraphics[width=0.43\textwidth,height=6.5cm]{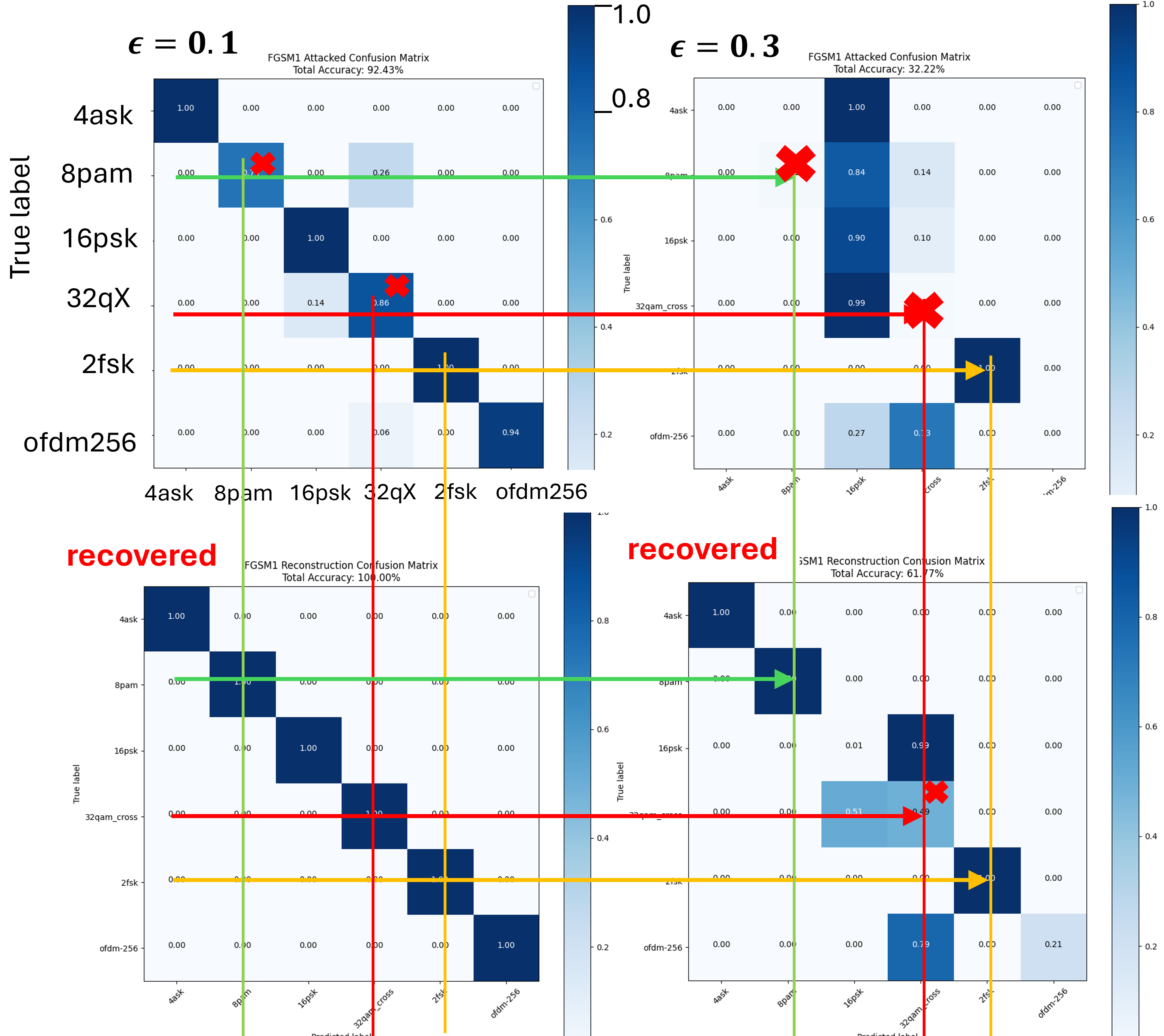} 
\caption{{\bf Confusion matrices after FGSM1, and upon passing through VQVAE: green arrow indicate PAM, red 32-QAM-X and yellow 2FSK, the 3 classes featured in histograms.}}\vspace{-3mm}
\label{fig:confFGSM1}
\end{figure}
%%%%%%%%%%%%%%%%%%%%%%%%%%%%%%%%%%
The degree in which accuracy changes with increasing attack strength clearly depends on modulation (Fig.~\ref{fig:confFGSM1}). The accuracy recovered by VQVAE processing also depends on the modulation. From Fig.~\ref{fig:confFGSM1}, following the yellow lines, the 2FSK classification seems to be very resilient to the attack. In communication settings, 2FSK is resilient to noise, having only two distant constellation points to distinguish between. However, spectrum sensing is different: instead of aiming to identify constellation points at the samples on symbol boundaries, a classifier looks for  the signatures expressed by all the samples. Hence, this classification robustness probably comes from the fact that 2FSK is one of the 2 constant envelope modulations (i.e. amplitude does not vary) in our dataset, hence easy to distinguish from other modulations even in the presence of a 30\% perturbation - see Fig.~\ref{fig:constells}. Despite this robustness, we see from Fig.~\ref{fig:Distances} that the discrete latents of 2FSK have the highest degree of change with $\epsilon$. A possible explanation is that 2FSK is oversampled at 8 IQ samples per symbol while all other datapoints have 2 samples per symbol. This creates a larger attack space per symbol, and causes high variability of datapoints. Focusing on FGSM1 attack in Fig.~\ref{fig:Distances}, we see that while the Set distance under $\epsilon=0.3$ is highest for 2FSK, the Hamming distance (2nd row) is the lowest of all modulations. We should now recall that $\Omega_{SC}$ performed by our VQVAE model is stochastic: {\em we do not quantize to the closest codeword}, but instead we sample from the learned distribution of closest codewords (Eqn.~\eqnref{post}). The low Hamming distance means that not many tokens in the "attacked latent" $Z_q^{af1}$ changed with respect to any single instantiation $\mathbf{z_q}^{(i)}$ of the non-attacked latent $Z_q$. The high Set distance means that many novel indices were sampled by $\Omega_{SC}$ to replace those tokens, i.e. from outside of the set of indices present in $Z_q.$ This phenomenon is obvious in the bottom histogram of Fig.~\ref{fig:FSKhist} where we see some probability mass in the range of 'insignificant codewords' (codewords outside of support of the original latents, the empty spans marked by red arrows in Fig.~\ref{fig:FSKhist}), suggesting the presence of noise instead of salient information.

\begin{figure}[h]
\centering
\begin{tabular}{c}   
 %\\\vspace{-2mm}
 %\hspace{-3mm}  
 \hspace{-3mm} 
 32-QAM-X $\epsilon \in \curlb{0.01, 0.1, 0.3}$\\
 \includegraphics[width=0.45\textwidth,height=7.0cm]{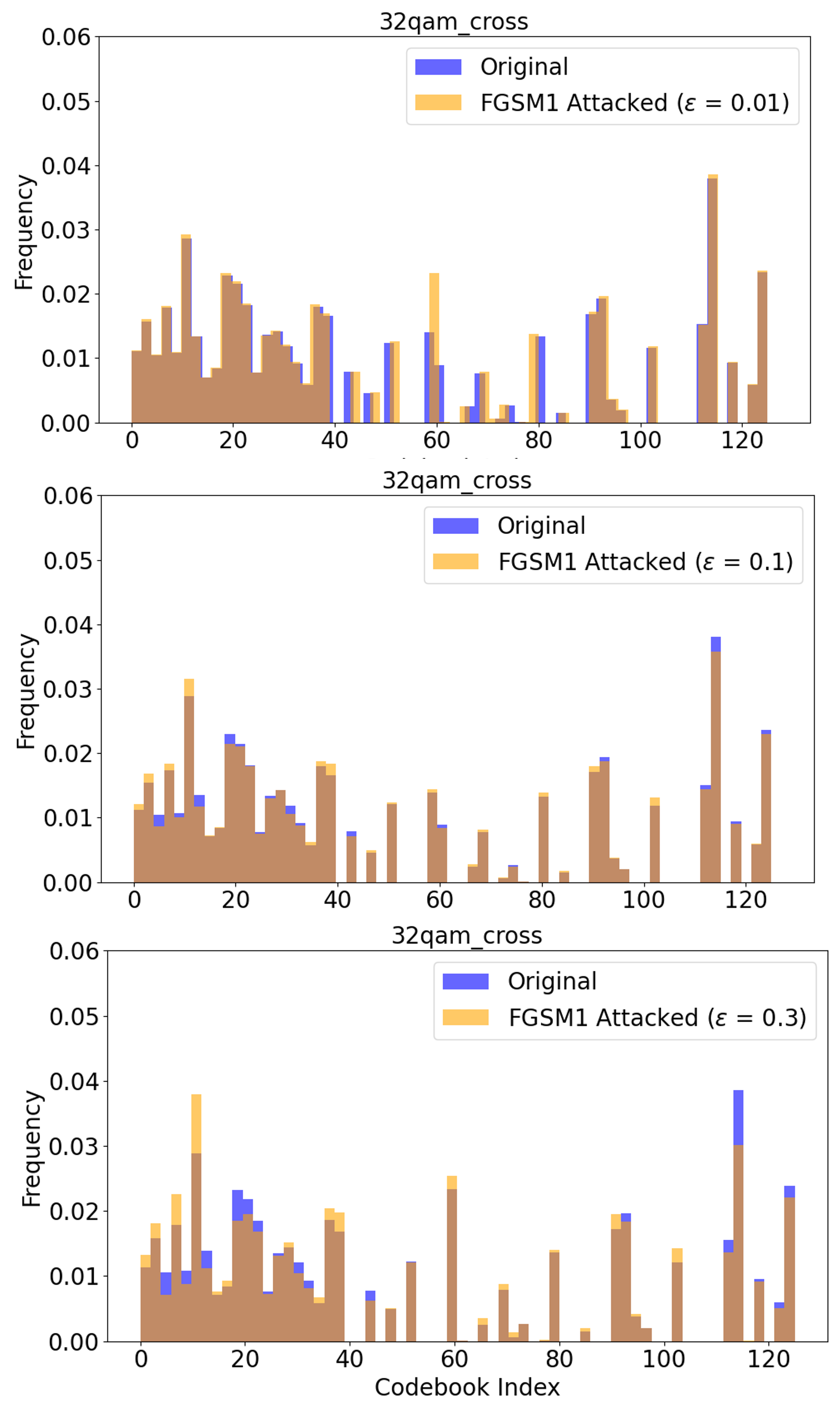} \\
 8PAM $\epsilon = 0.3$\\
\includegraphics[width=0.45\textwidth,height=2.33cm]{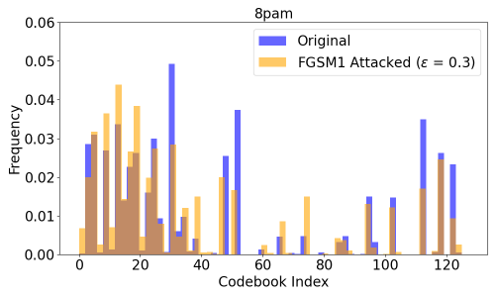}
\end{tabular}
\caption{{\bf Histograms of codeword indices under FGSM1 attacks for 32-QAM-X with $\epsilon \in \curlb{0.01, 0.1, 0.3}$ and 8PAM histogram with $\epsilon = 0.3$.}} \vspace{-6mm} 
\label{fig:32QAMhist}
\end{figure}

We here conjecture the following: as the codebook $Q$ is common for all waveform types, its dimension $q_s$ is much higher than the number of significant components in low-order modulations (e.g., 2FSK), allowing the attack to sway the $\Omega_{SC}$ mapping toward "`redundant"' codebook indices that are outside the support of this modulation's latent space. It has been shown \cite{dimred} that removing redundant features mitigates the attack, which is why autoencoders are good defense mechanisms\cite{AEagainst}.
%%%%%%%%%%%%%%%%%%%%%%%%%%%%%%%%%%%%%%%
%%%%%%%%%%%%%%%%%%%%%%%%%%%%%%%%%%%%%%%%%
%\subsection{Qualitative Evaluation}
Next, we picked a few  modulations to visualize the impact of the attack on their I/Q planes and how this impact is mitigated by the VQVAE, and we correlate it with  the shift in the codebook usage due to attacks. Apart from 2FSK, we pick one of the amplitude modulations (8 PAM), and  the 32-QAM-Cross (32-QAM-X), since it remains misclassified for $\epsilon=0.3$ even after the VQVAE (follow the red lines in Fig.~\ref{fig:confFGSM1}). 
\begin{itemize}
    \item \textbf{I/Q plane visualization:} After presenting ideal constellations of FSK, PAM and 32-QAM-X, Fig.~\ref{fig:constells} visualizes the IQ plane of a single random datapoint $x$ per class, its attacked version $x_a$ and $\hat{x}_a,$ the VQVAE reconstruction. 
    \item \textbf{Codebook Usage:} We present the codebook usage with the FGSM1-attacked latents for 2FSK and 32-QAM-X under 3 values of $\epsilon$  in Figs.~\ref{fig:FSKhist}~and~ \ref{fig:32QAMhist}, respectively. Because both $d_H$ and $d_S$ are very high for 8PAM under the FGSM1 attack (Fig.~\ref{fig:Distances}), we also include 8PAM for $\epsilon=0.3$ only (bottom of Fig~\ref{fig:32QAMhist}).
\end{itemize}
%%%%%%%%%%%%%%%%%%%%%%%%%%%%%%%%%%
%%%%%%%%%%%%%%%%%%%%%%%%%%%%%%%%%%%%%%%%%%%%%%%%%%%%%%%%%%%%%%%%%%%%%%%%%
Focusing on 32-QAM-X, which is a QAM modulation of unusual shape, observe from Fig.~\ref{fig:confFGSM1} that for $\epsilon=0.3$ it will not be recovered by VQVAE. The same holds  for the other 2 phase/QAM modulations, as opposed to the amplitude modulations, even though FGSM1 targets the amplitudes. The same repeats with FGSM2 at $\epsilon=0.3,$ only OFDM is recovering better. All data is completely irrecoverable for PGD2 at $\epsilon=0.3,$ except for 2FSK.
%%%%%%%%%%%%%%%%%%%%%%%%%%%%%%%%%%

On the other hand, for $\epsilon=0.2$ in all 3 rows in Fig.~\ref{fig:constells}, the last column exhibits similarity with the first one, meaning that the constellation shape gets recovered by VQVAE fairly well at this $\epsilon$, as evident by the accuracy plot in Fig.~\ref{fig:accuracy}.
Please observe that the attack $\epsilon=0.3$ is unreasonable and easily detectable by multiple tests. We are considering it only to gain insight how the latent indices and VQVAE are behaving, and how the I/Q constellations are changing in the limit of $SNR_a=20log_{10}\frac{std(X)}{\epsilon}\rightarrow 0$. There seem to be an inflection point after $\epsilon=0.2$ in which VQVAE starts experiencing a mode collapse, likely because
$std(X_{train})=0.25,$ meaning $SNR_a\approx 0$.
%%%%%%%%%%%e%%%%%%%%%%%%%%%%%%%
\vspace{-2mm}
\section{Conclusion}
We demonstrated the ability of VQVAE to suppress the strength of the adversarial attack on RF communication waveforms by analyzing waveform classification accuracy, I/Q diagrams, and various metrics derived from the discrete space learned by the model. Our findings reveal that both the susceptibility to attacks and the effectiveness of suppression vary with modulation type and sampling rate. The impact of VQVAE's codebook size on the suppression also differs across waveform classes. Although the presented results primarily highlighted the attack (FGSM1) whose intention is to target amplitude modulations more, it is difficult to establish its increased effect on those modulations, due to the impact of all other factors. Hence, attacks on modulation classifiers heavily depend on the subset of modulation classes used for training. In future work, we aim to develop a multi-feature attack detector, which will leverage our findings about the discrete space and the statistics from the waveform space.
\vspace{-2mm}
%%%%%%%%%%%%%%%%%%%%%%%%%%%
\bibliographystyle{IEEEtran}
\bibliography{advbib}
\end{document}